\documentclass[sigconf]{acmart}

\AtBeginDocument{%
  \providecommand\BibTeX{{%
    \normalfont B\kern-0.5em{\scshape i\kern-0.25em b}\kern-0.8em\TeX}}}

\copyrightyear{2022}
\acmYear{2022}
\setcopyright{acmcopyright}
\acmConference[AIES'22]{Proceedings of the 2022 AAAI/ACM Conference on AI, Ethics, and Society}{August 1--3, 2022}{Oxford, United Kingdom}
\acmBooktitle{Proceedings of the 2022 AAAI/ACM Conference on AI, Ethics, and Society (AIES'22), August 1--3, 2022, Oxford, United Kingdom}
\acmPrice{15.00}
\acmDOI{10.1145/3514094.3534159}
\acmISBN{978-1-4503-9247-1/22/08}

\usepackage{multirow}
\usepackage{subcaption}
\usepackage{dsfont} 
\usepackage{todonotes} 

\begin{document}

\title{Fairness via Explanation Quality: \\Evaluating Disparities in the Quality of Post hoc Explanations}

\author{Jessica Dai}
\affiliation{%
  \institution{Brown University}
  \city{Providence, RI}
  \country{USA}
}
\email{jessica.dai@alumni.brown.edu}

\author{Sohini Upadhyay}
\affiliation{%
  \institution{Harvard University}
  \city{Cambridge, MA}
  \country{USA}
}
\email{supadhyay@g.harvard.edu}

\author{Ulrich Aïvodji}
\affiliation{%
  \institution{Université du Québec à Montréal}
  \city{Montréal, QC}
  \country{CA}
}
\email{aivodji.ulrich@uqam.ca}

\author{Stephen H. Bach}
\affiliation{%
  \institution{Brown University}
  \city{Providence, RI}
  \country{USA}
}
\email{stephen_bach@brown.edu}

\author{Himabindu Lakkaraju}
\affiliation{%
  \institution{Harvard University}
  \city{Cambridge, MA}
  \country{USA}
}
\email{hlakkaraju@hbs.edu}

\newcommand{\jessica}[1]{{\color{orange}Jess: #1}}
\newcommand{\steve}[1]{{\color{blue}Steve: #1}}
\newcommand{\sohini}[1]{{\color{cyan}Sohini: #1}}
\newcommand{\hima}[1]{{\color{red}Hima: #1}}
\newcommand{\ulrich}[1]{{\color{green}Ulrich: #1}}
\newcommand{\hideh}[1]{}

\begin{abstract}
As post hoc explanation methods are increasingly being leveraged to explain complex models in high-stakes settings, it becomes critical to ensure that the quality of the resulting explanations is consistently high across all subgroups of a population. 
For instance, it should not be the case that explanations associated with instances belonging to, e.g., women, are less accurate than those associated with other genders. 
In this work, we initiate the study of identifying group-based disparities in explanation quality. 
To this end, we 
first outline several key properties that contribute to explanation quality---namely, fidelity (accuracy), stability, consistency, and sparsity---and discuss why and how disparities in these properties can be particularly problematic. 
We then 
propose an evaluation framework which can quantitatively measure disparities in the quality of explanations.
Using this framework, we carry out an empirical analysis with three datasets, six post hoc explanation methods, and different model classes 
to understand if and when group-based disparities in explanation quality arise. Our results indicate that such disparities are more likely to occur when the models being explained are complex and non-linear. We also observe that certain post hoc explanation methods 
(e.g., Integrated Gradients, SHAP) 
are more likely to exhibit 
disparities. 
Our work sheds light on previously unexplored ways in which explanation methods may introduce unfairness in real world decision making. 
\end{abstract}


\maketitle

\section{Introduction}
\label{sec:intro}
\begin{figure*}[ht!]
\centering 
\includegraphics[width=5in]{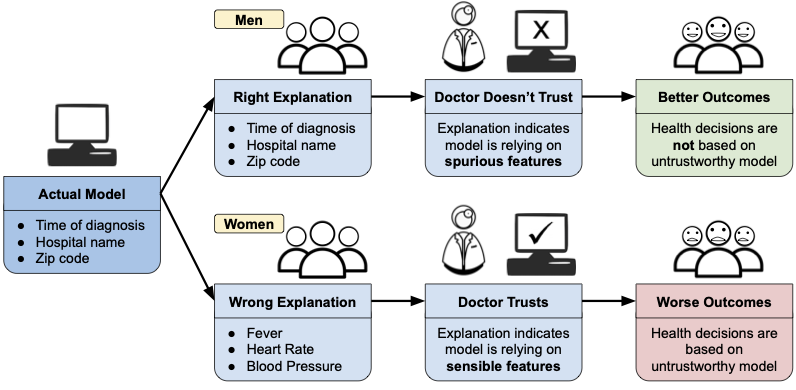}
\caption{
Illustrative example highlighting how disparities in explanation quality might lead to sub-optimal decisions, in turn resulting in worse real world outcomes for specific subgroups. 
}
\vspace{-0.2in}
\label{fig1}
\end{figure*} 
As machine learning (ML) models are increasingly being deployed to make consequential decisions in domains such as healthcare, finance, and policy, there is a growing
need to ensure
interpretable to ML practitioners and other domain experts (e.g., doctors, policy makers). Only if these practitioners have a clear picture of the behavior of these models can they assess when and how much to rely on them, and detect systematic errors and potential biases in them~\cite{doshi2017towards}. However, the increasing complexity as well as the proprietary nature of predictive models make it challenging to understand these complex black boxes, thus motivating the need for tools 
that can explain them in a faithful and human interpretable manner. To this end, several techniques have been proposed 
to explain complex models in a \emph{post hoc} fashion. 

Owing to their generality, post hoc explanation methods are increasingly being used to explain a number of complex models in high stakes domains such as medicine, finance, law, and science~\cite{elshawi2019interpretability,ibrahim2019global,whitmore2016mapping}. Therefore, it becomes critical to ensure that the explanations generated by these methods are of high quality. Prior research has studied several notions of explanation quality, including fidelity, stability, consistency, and sparsity~\cite{liu2021synthetic,petsiuk2018rise,slack2021reliable,zhou2021evaluating}. In this work, we focus on disparities in the \textit{quality} of explanations corresponding to different population subgroups.

To illustrate, consider a healthcare setting in which a predictive model is being used to aid doctors in diagnosing diseases (Figure~\ref{fig1}). 
Suppose
the model incorrectly relies on spurious features (e.g., hospital name, zip code) to make predictions. 
Let us assume that post hoc explanations generated by a state-of-the-art method are being provided to doctors to explain each model prediction. 
Further, explanations corresponding to 
men 
are accurate, and rightly capture that the underlying model is relying on spurious features. 
Therefore, doctors who examine these explanations are likely to mistrust the model in the case of male patients and make better decisions by employing their own discretion.
On the other hand, explanations 
for women 
are comparatively less accurate, and are incorrectly suggesting that the underlying model is relying on relevant features (e.g., fever, heart rate).  So, doctors may trust model predictions in the case of female patients due to the misleading explanations, and may therefore make incorrect decisions leading to unfairness (women are more likely to be incorrectly diagnosed than men) in the overall decision making process. 

The above scenario, while stylized, demonstrates how disparities in the quality of explanations may induce unfairness in real world decision making, and negatively impact the outcomes of vulnerable populations. Therefore, it is important to assess the extent to which such group-based disparities occur in the quality of the explanations output by state-of-the-art methods. However, there is little to no research that focuses on this critical problem. 

In this work, we 
initiate 
the study of group-based disparities in explanation quality. 
More specifically, we make the following key contributions: 
\begin{itemize}
    \item We formulate the problem of detecting group-based disparities in explanation quality.
    \item We propose a novel evaluation framework which can quantitatively measure disparities in the quality of explanations output by state-of-the-art methods.
    We show an example of how to use this framework and some initial directions for inquiry using existing common metrics---specifically, \emph{fidelity disparity}, \emph{stability disparity}, \emph{consistency disparity}, and \emph{sparsity disparity}. 
    \item We leverage the aforementioned framework to carry out a rigorous empirical analysis with six state-of-the-art post hoc explanation methods, three real world datasets, and two different model classes (linear models and deep neural networks) to study if and when group-based disparities in explanation quality arise. 
\end{itemize}
Results from our empirical analysis indicate that disparities in explanation quality are 
more likely to occur when the models being explained are complex and highly non-linear. 
Furthermore, we also observed that post hoc explanation methods such as Integrated Gradients and SHAP are more likely to generate explanations that are prone to the aforementioned disparities. 
Our analysis also revealed that group-based disparities are most prominent in case of two notions of explanation quality---namely, fidelity and sparsity. Our findings not only shed light on the previously unexplored problem of group-based disparities in explanation quality, but also pave the way for rethinking the design and development of explanation methods in a way that such disparities can be minimized.

\section{Related Work} 
\label{sec:related}

This work builds on extensive work in explainable machine learning. 
Crucially, the usage of explanations is often motivated by higher-stakes settings, where the interpretation of the explanation has influence on how a human chooses to interact with the model. Common use-cases for explanations across a variety of stakeholders include 
model debugging and improvement, monitoring, building confidence, transparency, and auditing
\citep{bhatt2020explainable, hong2020human, suresh2021beyond}. 
Most notably, explanations can be used for informing and justifying downstream actions for decision-making. 
For instance, based on the contents of the explanation, an individual might decide to  follow the model's recommendation (or not); or, in a different application setting, an individual might decide to contest the result of a model's decision (or not). 
The related literature described in this section ultimately works towards these broader goals of \textit{usefulness} for explanations, and it is these broader goals that our concerns about performance disparity ultimately address.

\vspace{4pt}\noindent \textbf{Inherently Interpretable Models and Post hoc Explanations}
Many approaches learn inherently interpretable models such as rule lists \citep{zeng2017interpretable, wang2015falling}, decision trees and decision lists \citep{letham2015interpretable}, 
and others 
\citep{lakkaraju2016interpretable, bien2009classification, lou2012intelligible, caruana15:intelligible}.
However, complex models such as deep neural networks often achieve higher accuracy than simpler  models~\cite{lime}.
Thus, there has been significant interest in constructing post hoc explanations to understand their behavior. To this end, several techniques have been proposed in recent literature to construct \emph{post hoc explanations} of complex decision models.
For instance, LIME, SHAP, Anchors, BayesLIME, and BayesSHAP~\cite{lime,shap,ribeiro2018anchors,slack2021reliable} are considered \emph{perturbation-based local} explanation methods because they leverage perturbations of individual instances to construct interpretable local approximations (e.g., linear models).
On the other hand, methods such as Gradient $^*$ Input, SmoothGrad, Integrated Gradients, and GradCAM~\cite{simonyan2013saliency, sundararajan2017axiomatic, selvaraju2017grad,smilkov2017smoothgrad} are referred to as \emph{gradient-based local} explanation methods since they leverage gradients computed with respect to input features of individual instances to explain individual model predictions. An alternate class of methods referred to as \emph{global} explanation methods attempt to summarize the behavior of black-box models as a whole rather than in relation to individual data points \citep{lakkaraju2019faithful,bastani2017interpretability}. A more detailed treatment of this topic is provided in other comprehensive survey articles~\citep{arrieta2020explainable, guidotti2018survey, murdoch2019definitions, linardatos2021explainable,covert2021explaining}. In contrast to the aforementioned works, which propose novel methods to explain models and their predictions, our work focuses on understanding group-based disparities in the quality of explanations generated by state-of-the-art post hoc local explanation methods. 

\vspace{4pt}\noindent \textbf{Evaluating Explanations}
Prior research has proposed several metrics to determine if an explanation is reliable. 
An extensive survey of metrics for evaluating explanation methods can be found in \citet{zhou2021evaluating}, who propose two high-level goals of explanation methods: interpretability (the clarity, simplicity, and broadness of the explanations), 
and fidelity (the completeness and soundness of explanations)~\cite{carvalho2019machine, gilpin2018explaining}. 
\citet{liu2021synthetic} provide a synthetic benchmark for explanation evaluation which includes implementations of several metrics, as well as a discussion of how to choose metrics for evaluation. Furthermore, several prior works proposed different metrics for evaluating various aspects of explanation quality, such as fidelity, stability, consistency, and sparsity~\cite{liu2021synthetic,petsiuk2018rise,slack2021reliable,zhou2021evaluating,ghorbani2019interpretation,alvarez2018robustness, hooker2018evaluating, lakkaraju2019faithful}. 
Recent research further leveraged the aforementioned properties and metrics to theoretically and empirically analyze the behavior of popular post hoc explanation methods~\cite{ghorbani2019interpretation, slack2019can,dombrowski2019explanations,adebayo2018sanity,alvarez2018robustness,levine2019certifiably,pmlr-v119-chalasani20a,agarwal2021towards}. 

In addition to quantitative metrics,
\textit{human-grounded} evaluation approaches
emphasize how human users perceive and utilize explanations \citep{doshi2017towards}. 
For example, \citet{lakkaraju2020fool} carry out a user study to understand if misleading explanations can fool domain experts into deploying racially biased models, while \citet{kaur2020interpreting} find that explanations are often over-trusted and misused. 
Similarly, \citet{poursabzi2018manipulating} find that supposedly-interpretable models can lead to a decreased ability to detect and correct model mistakes, possibly due to information overload.
\citet{jesus2021can} introduce a method to compare explanation methods based on how subject matter experts perform on specific tasks with the help of explanations.
\citet{lage2019evaluation} use insights from rigorous human-subject experiments to inform regularizers used in explanation algorithms. 
Though our work does not involve human subject study, we see this line of human-centered investigation as a critical complement to our quantitative evaluation framework.

\vspace{4pt}\noindent\textbf{Limitations and Vulnerabilities of Post hoc Explanations}
The aforementioned notions of explanation quality and the corresponding evaluation metrics were also leveraged to analyze the behavior of post hoc explanation methods and their vulnerabilities---e.g.,~\citet{ghorbani2019interpretation} and~\citet{slack2020fooling} demonstrated that methods such as LIME and SHAP may result in explanations that are not only inconsistent and unstable, but also prone to adversarial attacks.
Furthermore, ~\citet{lakkaraju2020fool} and~\citet{slack2020fooling} showed that explanations which do not accurately represent the importance of sensitive attributes (e.g., race, gender) could potentially mislead end users into believing that the underlying models are fair when they are not~\cite{lakkaraju2020fool,slack2020fooling,aivodji2019fairwashing}. This, in turn, could lead to the deployment of unfair models in critical real world applications.
There is also some discussion about whether models which are not inherently interpretable ought to be used in high-stakes decisions at all.
\citet{rudin2019stop} argues that post hoc explanations tend to be unfaithful to the model to the extent that their usefulness is severely compromised. While this line of work demonstrate different ways in which explanations could potentially induce inaccuracies and biases in real world applications, they do not focus on analyzing group-based disparities in explanation quality, which is our focus. 

\vspace{4pt}\noindent\textbf{The Intersection of Fairness and Explainability} 
While fairness and explainability are often described as qualities which ``responsible'' models ought to have simultaneously \citep{toreini2020relationship, schumann2020we}, there is surprisingly little work exploring the intersection of the two. 
\citet{begley2020explainability} extend SHAP as a tool to explain observed (un)fairness. Several recent works \citep{abdollahi2018transparency, stevens2020explainability, soares2019fair, ortega2021symbolic} 
propose methods to make \textit{specific} algorithms simultaneously fair and explainable. As fairness is often explicitly referred to as a justification or motivation for the use of explanation methods
\citep{suresh2021beyond, bhatt2020machine, chen2018fair}, it is critical to also \textit{evaluate} explanations from the perspective of fairness.
Here, we make a distinction between
whether an explanation \textit{accurately portrays the fairness of the underlying model}, and whether the explanation method \textit{performs equally well on all groups}. 
Several works address the former question---e.g.,
\citet{aivodji2019fairwashing} introduce the notion of ``fairwashing,'' in which a rule-based explanation method can be adversarially constructed such that an unfair model decision can be rationalized. 
Meanwhile, \citet{slack2020fooling} illustrate how a model can be adversarially constructed such that LIME and SHAP hide
the explicitly unfair behavior of the model. \citet{alikhademi2021can} finds that existing explainability tools are ill-suited for evaluating fairness performance. 
Finally, \citet{dodge2019explaining} conduct a human study investigating how explanations can be used to interpret the fairness performance of a model.
While the aforementioned works address the question of whether an explanation \emph{accurately portrays the fairness of the underlying model}, we investigate whether the explanation method \emph{performs equally well on all the groups}.
In concurrent work,  \citet{balagopalan2022road} are motivated by a similar question. Notably, they focus solely on fidelity, and empirically validate our findings of disparity in existing explanation methods.

\section{Our Framework}
\label{sec:background}

\subsection{Preliminaries and Notation}
Let us consider a dataset $\mathcal D = \{(\mathbf{x}^{(1)}, y^{(1)}), \dots, (\mathbf{x}^{(n)}, y^{(n)})\}$ where $\mathbf{x}^{(i)} \in \mathbb{R}^d$ is a vector of feature values and represents an instance in the data, and $y^{(i)} \in \mathcal{Y}$ is the corresponding class label. Since our goal is to unearth group-based disparities in the quality of post hoc explanations, we consider the case where the dataset $\mathcal{D}$ comprises of some feature $s$ which corresponds to a discrete sensitive attribute (e.g., gender, race, age).

Let $f:\mathbb{R}^d\to\mathcal{Y}$ be a predictive model which can make predictions on instances in $\mathcal{D}$. 
Let $\mathcal{E}:(\mathbf{x}, f)\mapsto \mathbf{w} \in \mathbb{R}^d$ be a local explanation method which takes as input an instance $\mathbf{x}$ and the predictive model $f$, and outputs a vector of feature importances $\mathbf{w}$. Note that the $j$-th element of this vector (i.e., $w_j$) represents how important the $j$-th feature
is to the prediction $f(\mathbf{x})$. 
Finally, let $\mathcal{M}:(\mathbf{x}, f,  \mathcal{E})\mapsto \mathbb{R}$ denote a metric which measures some facet of the quality (e.g., fidelity or stability etc.) of a given local explanation $\mathcal{E}(\mathbf{x}, f)$. 

\subsection{Metrics for Evaluating Explanations}

In their survey,
\citet{zhou2021evaluating} highlight two characteristics of a high-quality explanation: first, how well the explanation approximates the model, and second,  human-understandability of the explanation.
We focus on three metrics---fidelity, stability, and consistency---that measure the accuracy of the explanation's approximation, as well as 
a fourth---sparsity---that measures how easily the explanation is understood by a user. 

\subsubsection{Fidelity}
Perhaps the most natural question when considering explanation quality is the extent to which it accurately represents the underlying decision-making process.
In other words, are the ``important features'' designated in the explanation \textit{truly} important features for the model?
Explanations that accurately designate the important features to the underlying model are said to have high \emph{fidelity}.

In our framework, we select two measures of fidelity: \emph{ground truth fidelity}, which is applicable to models that represent their features as linear weights comparable with explanations, and \emph{prediction gap fidelity}, which is applicable to a wider range of models like neural networks.

The ground truth fidelity metric applies solely to machine learning models which inherently encode some notion of feature importance as a vector of $d$ weights, 
such as in the coefficients of a logistic regression model.
While therefore limited in practical applicability across most models, this metric is a useful baseline for determining
whether an explanation method is correct when there exists a ground truth definition of features and their important.
Specifically, suppose we have ground truth weights $\mathbf{\omega}$, such as the weights of a logisitic regression model.
The ground truth fidelity metric is defined as the percent of the top $k$ features from explanation $\mathbf{w}$ which are also in the top $k$ features from $\mathbf{\omega}$.

\begin{definition}
Let $\mathbf{\omega} \in \mathbb{R}^d$ be a vector denoting the importance of features in a model $f$, where $\omega_i > \omega_{i^\prime}$ implies that the $i$th feature is more important in $f$ than the $i^\prime$th feature.
Let $\text{top}(k, \cdot)$ be a function that returns the indices of the $k$ largest elements of a given input vector.
Then, given local explanation $\mathbf{w}=\mathcal{E}(\mathbf{x},f)$ the \textbf{ground truth fidelity} for a given value of $k$ is defined as follows:
\begin{equation*}
    \mathcal M_\text{ground truth}(\mathbf{x}, f, \mathcal{E}) = 
    \frac{|\text{top}(k, \mathbf{w}) \cap \text{top}(k, \mathbf{\omega})|}{k} \; .
\end{equation*}
\end{definition}

The above definition requires that some agreed-upon measure of feature importance $\mathbf{\omega}$ is available.
For models that do not have such a quantity, such as deep neural networks, we measure prediction gap fidelity.
The intuition behind this fidelity metric is that the model's prediction \textit{should not} change substantially if minor alterations are made to features that are \textit{not} important. 
If the prediction does change substantially, then this indicates that the explanation failed to capture features that were in fact important to the prediction, and is thus low fidelity.
This follows a similar intuition to saliency in \citet{dabkowski2017real}, and insertion/deletion in \citet{petsiuk2018rise}. 
This metric is applicable for probabilistic classifiers $f(\mathbf{x}) = g(h(\mathbf{x}))$, where $h: \mathbb{R}^d \rightarrow [0,1]$ generates predicted probabilities and $g: [0,1] \rightarrow \{0, 1\}$ maps the probability to a classification. The $k$ most important features of the model for $\mathbf{x}$ are identified from $\mathbf{w} = \mathcal{E}(\mathbf{x}, f)$. Then, a small amount of Gaussian noise with variance $\sigma$ is added to all features of $\mathbf{x}$ outside the top $k$ to produce $\tilde{\mathbf{x}}$, 
and the difference between $h(\mathbf{x})$ and $h(\tilde{\mathbf{x}})$ is taken. 
The ``prediction gap'' for the single $\mathbf{x}$ is the expected value of this difference;
practically, the procedure of adding noise and recording the gap in prediction---is repeated a total of $m$ times, and the ``prediction gap'' is computed as the average of all $m$ gaps. 

\begin{definition}
Let $f$ be a model such that $f(\mathbf{x}) = g(h(\mathbf{x}))$, where $h: \mathcal{X}\rightarrow [0,1]$ generates predicted probabilities and $g: [0,1] \rightarrow \{0, 1\}$ maps the probability to a classification.
Let $\text{top}(k, \cdot)$ be a function that returns the indices of the $k$ largest elements of a given input vector.
Then, given local explanation $\mathbf{w} = \mathcal{E}(\mathbf{x},f)$ and variance parameter $\sigma$, define $\tilde{\mathbf{x}}$ such that
\[
\tilde{x}_{i} = x_{i} + \mathds 1 [i \notin \text{top}(k, \mathbf{w})] \cdot z_{i}
\text{\;\;\;\; where \;\;\;\;} z_{i} \sim \mathcal N(0, \sigma)\; .
\]
Then {\bf prediction gap fidelity} is defined as follows:
\begin{equation*}
           \mathcal M_\text{pred. gap} (\mathbf{x}, f, \mathcal{E})
       = \mathbb E [|h(\mathbf{x)} - h(\tilde{\mathbf{x}})|] \; .
\end{equation*}
We approximate prediction gap fidelity empirically. We generate $\tilde{\mathbf{x}}$ $m$ times, producing $\{\tilde{\mathbf{x}}_{j} \mid 1\leq j \leq m\}$.
Then predication gap fidelity is approximated as follows:
\begin{equation*}
           \hat{\mathcal{M}}_\text{pred. gap} (\mathbf{x},f, \mathcal{E})
       = \frac{1}{m} \sum_{j=1}^m
       [|h(\mathbf{x)} - h(\tilde{\mathbf{x}}_{j})|] \; .
\end{equation*}
\end{definition}

\subsubsection{Stability}

The idea that similar points should receive similar explanations---and the observation that popular explanations often do not satisfy this property---has been discussed extensively (common terms for this principle, in addition to stability, include robustness and insensitivity) \citep{alvarez2018towards,bhatt2020evaluating, yeh2019fidelity}. 
\citet{alvarez2018towards} note that instability has been observed even when the underlying model is stable---calling into question the veracity of those explanations.
In other words, given a set of differing explanations for similar points, it is not credible that those explanations are all correct.

To measure an explanation's stability at a point $\mathbf{x}$, we add some noise to $\mathbf{x}$ to generate similar points, calculate explanations for those similar points, then take the average L1 distance between $\mathbf{x}$'s explanation and those of the similar points.
This is similar to the definition of prediction gap fidelity, except now we add noise to all features and measure the difference in explanations rather than model predictions.
This definition exactly mirrors the definition of stability in \citet{yeh2019fidelity} and others \citep{alvarez2018robustness, bhatt2020evaluating}.

\begin{definition}
Given a model $f$, local explanation ${\mathbf w} = \mathcal{E}({\mathbf x}, f)$, and variance parameter $\sigma$,
define $\tilde{\mathbf{x}}$ such that
\[
\tilde{x}_{i} = x_{i} + z_{i}
\text{\;\;\;\; where \;\;\;\;} z_{i} \sim \mathcal N(0, \sigma)\; .
\]
Then {\bf instability} is defined as follows:
\begin{equation*}
\mathcal M_\text{instability}(\mathbf{x}, f, \mathcal{E}) = \mathbb E \left[ \left\|\mathcal{E}(\mathbf{x},f) - \mathcal{E}(\tilde{\mathbf{x}},f) \right\|_1 \right] \; .
\end{equation*}

We approximate instability empirically.
We generate $\tilde{\mathbf{x}}$ $m$ times, producing $\{\tilde{\mathbf{x}}_{j} \mid 1\leq j \leq m\}$.
Then instability is approximated as follows:
\begin{equation*}
    \hat{\mathcal{M}}_\text{instability}(\mathbf{x}, f, \mathcal{E}) = \frac{1}{m}\sum_{j=1}^m
    \left\|\mathcal{E}(\mathbf{x},f) - \mathcal{E}(\tilde{\mathbf{x}}_j, f) \right\|_1 \; .
\end{equation*}
\end{definition}

\subsubsection{Consistency}

Consistency captures the intuition that if an explanation for a single point is calculated multiple times, each of the calculated explanations should be similar.
Inconsistent explanations for the same input $\mathbf{x}$ suggest that these explanations may be unreliable.
This metric is motivated by concerns raised by empirical observations of the performance of many stochastic post-hoc explanation methods---requesting many explanations for the same point often resulted in drastically different explanations \citep{slack2021reliable,alvarez2018robustness,lee2019developing,zhang2019should, lee2019developing}.
If the same point may result in a variety of very different explanations, it is unlikely that any individual explanation is correct, indicating a low approximation quality. 

To measure consistency for a point $\mathbf{x}$, we calculate several explanations for that same point; then, we take the average L1 distance between the first explanation and each of the new explanations.

\begin{definition}
Let an explanation method $\mathcal E$ be stochastic, such that $\mathcal E_j$ indicates an explanation generated with a random seed $j$. 
The empirical {\bf inconsistency} is defined as follows:
\begin{equation*}
    \hat{\mathcal {M}}_\text{inconsistency}(\mathbf{x},f,\mathcal{E}) = \frac{1}{m}\sum_{j=1}^m
    \left \|\mathcal E_0(\mathbf{x}, f) - \mathcal E_j(\mathbf{x}, f) \right\|_1 \; .
\end{equation*}
\end{definition}

\subsubsection{Sparsity}

Finally, we are also interested in how easily an explanation can be understood and interpreted by users. 
Extensive previous work drawing from cognitive psychology has discussed the question of whether a given explanation is actually understood by a human user \citep{wang2019designing, narayanan2018humans,kliegr2021review}.
One common theme is complexity leading to higher cognitive load.
Interpretability is measured in \citet{bhatt2020evaluating} as the entropy of the explanation, where equal attribution to all features is considered to be the least-interpretable.
In \citet{lakkaraju2019faithful}, interpretability is measured by counting discrete concepts.

We measure sparsity by counting the number of features with an attributed importance greater than a threshold $t$. 
\begin{definition}
Let $t$ be the threshold for which a feature importance is considered significant.
Given local explanation $\mathbf{w} = \mathcal{E}(\mathbf{x},f)$ and threshold parameter $t$, {\bf complexity} 
is defined as follows: 
\begin{equation*}
    \mathcal M_\text{complexity}(\mathbf{x}, f, \mathcal{E})
    = \sum_i^d \mathds 1 [w_i > t] \; .
\end{equation*}
\end{definition}


\subsection{Measuring Disparity}

For any given metric, we can empirically estimate disparity as follows. For each dataset, we partition the data into groups according to the sensitive attribute. In the datasets used for our experiments, we split the data into group 0 (majority group) and group 1 (minority group), though our framework is generic enough to handle multi-valued sensitive attributes. For each of these groups, we compute the mean values of the metric by averaging the metric values across instances in the respective groups. If there is a significant difference (as measured by a hypothesis test) between the mean metric values for each group, then there is a disparity in the quality of the explanations for that metric. 

\subsection{Consequences of Disparity}

Though our broad proposal is to measure disparities in explanation quality \textit{in general}, in this section we discuss some possible consequences of disparity in the specific metrics defined in the previous sections. We emphasize that disparity in additional metrics---or even different implementations of the same metrics---may have implications beyond what is discussed here; furthermore, this discussion is meant primarily as a preliminary exploration, and more tailored study (for example, with human subjects) for each case described here is warranted.

\subsubsection{Fidelity Disparity}

A significant disparity in explanation fidelity, or the \textit{closeness} of an explanation to the model's actual behavior, can have concrete ramifications.
For example, as in Figure~\ref{fig1}, a domain expert such as doctor could make systematically worse decisions for a group of people because of a disparity in explanation quality.

\subsubsection{Stability Disparity}

Like disparity in fidelity, disparity in stability can be consequential. 
Suppose the machine learning model that a doctor uses an explanation method that is not equally stable across groups.
This disparity might lead to them rely, rightly or wrongly, more often on the model being explained for one group or the other.
\textit{Measuring} disparity in stability at development time, meanwhile, is important not just to prevent the above scenario, but to potentially identify problems with the model itself.
\citet{dooley2021robustness} observed that models can be unstable across groups; disparity in explanation stability may be a warning sign.

\subsubsection{Consistency Disparity.}

Disparity in consistency is also problematic when considering the user's interpretation of the explanations.
A user may themselves be testing the explanation method by requesting an explanation multiple times, choosing to use the explanation for a given point only if they observe that the multiple explanations they received for that point are similar. 
If consistency disparity exists, therefore, the user may end up using explanations to aid decisionmaking much more often for one group than another.
Returning to the example of the clinical setting, this may ultimately result in meaningfully disparate diagnoses.

\subsubsection{Sparsity Disparity}

Unlike the three metrics discussed previously, disparity in sparsity across demographic groups is not inherently problematic.
In fact, it may reflect the model's true behavior, e.g., if the model has a more complex decision boundary for one group than another, or if the model does rely more features for one group than another. 
Checking for disparity in sparsity, therefore, can be informative about the underlying model. 
One possible adverse consequence of disparity in sparsity is that there may be higher cognitive load when examining explanations for one group than another. 
However, if it is the case that the model truly is relying on more information to make predictions for one group, 
artificially ensuring equal sparsity may result in a substantial tradeoff with fidelity.

\section{Empirical Analysis}
\label{sec:experiment}

In this section, we demonstrate how to employ our explanation evaluation framework and interpret its outputs. First we generate a range of post-hoc explanations for various data and model combinations. We apply our evaluation framework to these explanations, computing the metrics of interest and analyzing them across data subgroups. Finally we examine the disparities that emerge, highlighting trends that may be of interest to practitioners.
\subsection{Experimental Setup}
\subsubsection{Data} Our evaluation framework is of particular importance in settings where explanation disparities emerge across protected attribute classes like sex and race. Accordingly, we choose the following benchmark datasets: German Credit \cite{UCI}, Student Performance \cite{student,UCI}, and COMPAS \cite{compas}. 
Though we acknowledge the limitations of COMPAS as a benchmark dataset \citep{bao2021s} (and indeed, ``fairness benchmarks'' in general), we choose to evaluate and report explanation performance on it because we do not claim to successfully ``achieve'' any notion of fairness with respect to classification or explanation.
We evaluate explanation disparities across one protected attribute in each dataset, namely sex in both German Credit and Student Performance and race in COMPAS.
Datasets are divided into training and testing sets via an 80/20 random stratified split with respect to the target label.
\subsubsection{Models}
We investigate whether explanation disparities can vary based on the complexity of model being explained.
We train both linear and non-linear models, Logistic Regression (LR) and a 3-layer neural network (NN) respectively.
For the latter, we use 50, 100, and 200 nodes in each consecutive layer, ReLU activation, binary cross entropy loss, Adam optimizer, and 100 training epochs.
\subsubsection{Explanation Methods} 
We investigate whether different explanation methods can produce different disparities. Most post-hoc explanation methds can be categorized into two families of approaches. One approach generates explanations by constructing locally interpretable model approximations whereas the other generates explanations by analyzing the behavior of the model's gradient. We choose LIME \cite{lime}, SHAP \cite{shap}, and MAPLE \cite{maple} from the former and SmoothGrad \cite{smilkov2017smoothgrad}, Integrated Gradients (IntGrad) \cite{sundararajan2017axiomatic}, and Vanilla Gradients (VanGrad) \cite{simonyan2013saliency} from the latter. We employ the authors'  implementations for LIME \cite{lime} and MAPLE \cite{maple}. We use Captum library \cite{captum} implementations of SHAP (sample size 1000), IntGrad, and VanillaGrad. We use  default hyperparameter settings throughout unless otherwise specified. We implement SmoothGrad in PyTorch with standard normal noise and sample size 1000.

\subsubsection{Metric Hyperparameters} For both \textbf{ground truth fidelity} and \textbf{prediction gap fidelity} we fix $k=5$. Additionally, for \textbf{prediction gap fidelity} we fix $m=1000$ and $\sigma=0.1$. For both \textbf{instability} and \textbf{inconsistency} we fix $m=5$. For \textbf{sparsity} we fix $t=0.01$.

\subsubsection{Setting and Implementation Details}
For each dataset $\mathcal{D}$, we begin by splitting the data into training set $\mathcal{D}_{train}$ and testing set $\mathcal{D}_{test}$ with respect to a random seed, as detailed in 4.1.1. Next we train predictive model $f$ on $\mathcal{D}_{train}$ and partition $\mathcal{D}_{test}$ with respect to the value of the salient protected attribute into $\mathcal{D}_{0}$ and $\mathcal{D}_{1}$, where $\mathcal{D}_{0}$ and $\mathcal{D}_{1}$ correspond to elements in $\mathcal{D}_{test}$ with protected attribute values of $0$ and $1$ respectively. For each explanation method $\mathcal{E}$ and framework metric $\mathcal{M}$, we generate $M_{0} = \{\mathcal{M}(x,f, \mathcal{E}) \mid x \in \mathcal{D}_{0}\}$ and $M_{1} = \{\mathcal{M}(x,f, \mathcal{E}) \mid x \in \mathcal{D}_{1}\}$. By comparing $M_0$ and $M_1$, we can determine whether there is a disparity in explanation performance. To understand whether $M_0$ and $M_1$ consistently differ, we perform 5 trials, repeating this procedure with different random seeds for the data split.

\begin{table*}[ht!]
\small
\begin{subtable}[h]{\textwidth}
\centering
\begin{tabular}{|c|c|c|c|c|c|c|c|}
\hline
 &  & LIME & SHAP & SmoothGrad & IntGrad & VanillaGrad & Maple \\ \hline
\multirow{1}{*}{German Credit} & LR & 0.424 & \textbf{0.008} & 1.0 & \textbf{0.008} & 1.0 & 0.905 \\ \hline
\multirow{1}{*}{Student Performance} & LR & 1.0 & 0.421 & 1.0 & 0.421 & 1.0 & 1.0 \\ \hline
\multirow{1}{*}{COMPAS} & LR & 0.841 & 0.151 & 1.0 & 0.131 & 1.0 & 0.401 \\  \hline
\end{tabular}
\caption{Ground truth --- 2/18 significant}
\end{subtable}
\begin{subtable}[h]{\textwidth}
\centering
\begin{tabular}{|c|c|c|c|c|c|c|c|}
\hline
 &  & LIME & SHAP & SmoothGrad & IntGrad & VanillaGrad & Maple \\ \hline
\multirow{2}{*}{German Credit} & LR & \textbf{0.032} & 0.056 & \textbf{0.032} & 0.056 & \textbf{0.032} & 0.421 \\ \cline{2-8}
 & NN & 0.421 & 0.421 & 0.690 & 0.421 & 0.310 & 0.548 \\ \hline
\multirow{2}{*}{Student Performance} & LR & 0.691 & 0.548 & 0.690 & 0.549 & 0.690 & 0.690 \\ \cline{2-8} 
 & NN & 0.056 & \textbf{0.016} & 0.056 & \textbf{0.016} & 0.056 & \textbf{0.031} \\ \hline
\multirow{2}{*}{COMPAS} & LR & 0.222 & \textbf{0.008} & 0.151 & 0.310 & 0.151 & 0.548 \\ \cline{2-8} 
 & NN & 0.095 & \textbf{0.016} & \textbf{0.008} & \textbf{0.016} & \textbf{0.016} & 0.222 \\ \hline
\end{tabular}
\caption{Prediction Gap --- 11/36 significant}
\end{subtable}
\begin{subtable}[h]{\textwidth}
\centering
\begin{tabular}{|c|c|c|c|c|c|c|c|}
\hline
 &  & LIME & SHAP & SmoothGrad & IntGrad & VanillaGrad & Maple \\ \hline
\multirow{2}{*}{German Credit} & LR & 0.100 & \textbf{0.008} & 1.0 & \textbf{0.008} & 0.690 & 0.690 \\ \cline{2-8} 
 & NN & 0.421 & 0.222 & \textbf{0.016} & \textbf{0.008} & \textbf{0.016} & 0.675 \\ \hline
\multirow{2}{*}{Student Performance} & LR & 0.690 & 0.016 & 1.0 & \textbf{0.008} & 0.841 & 1.0 \\ \cline{2-8} 
 & NN & 0.690 & \textbf{0.016} & 0.917 & \textbf{0.008} & 0.100 & 1.0 \\ \hline
\multirow{2}{*}{COMPAS} & LR & \textbf{0.007} & \textbf{0.008} & 1.0 & \textbf{0.008} & 0.158 & 0.690 \\ \cline{2-8} 
 & NN & 0.310 & 0.151 & 1.0 & 0.222 & 0.310 & 0.548 \\ \hline
\end{tabular}
\caption{Sparsity --- 11/36 significant}
\end{subtable}
\begin{subtable}[h]{\textwidth}
\centering
\begin{tabular}{|c|c|c|c|c|c|c|c|}
\hline
\multicolumn{1}{|r|}{} &  & LIME & SHAP & SmoothGrad & IntGrad & VanillaGrad & Maple \\ \hline
\multirow{2}{*}{German Credit} & LR & 0.222 & 0.222 & 0.548 & 1.0 & \textbf{0.016} & 1.0 \\ \cline{2-8} 
 & NN & 0.690 & 0.100 & 0.056 & 0.310 & 0.100 & 0.841 \\ \hline
\multirow{2}{*}{Student Performance} & LR & 0.690 & 0.690 & 0.548 & 0.690 & 0.310 & 1.0 \\ \cline{2-8} 
 & NN & 0.310 & 0.310 & 0.690 & 0.056 & 0.056 & 0.841 \\ \hline
\multirow{2}{*}{COMPAS} & LR & 0.421 & 0.222 & 0.222 & 0.222 & \textbf{0.008} & 0.841 \\ \cline{2-8} 
 & NN & 0.310 & \textbf{0.008} & 0.100 & \textbf{0.008} & \textbf{0.008} & 0.690 \\ \hline
\end{tabular}
\caption{Stability --- 5/36 significant}
\end{subtable}
\begin{subtable}[h]{\textwidth}
\centering
\begin{tabular}{|c|c|c|c|c|c|c|c|}
\hline
\multicolumn{1}{|r|}{} &  & LIME & SHAP & SmoothGrad & IntGrad & VanillaGrad & Maple \\ \hline
\multirow{2}{*}{German Credit} & LR & \textbf{0.016} & 1.0 & 1.0 & 1.0 & 1.0 & 0.690 \\ \cline{2-8} 
 & NN & 0.548 & 1.0 & 1.0 & 0.841 & 1.0 & 1.0 \\ \hline
\multirow{2}{*}{Student Performance} & LR & 0.421 & 1.0 & 1.0 & 1.0 & 1.0 & 1.0 \\ \cline{2-8} 
 & NN & 0.690 & 0.548 & 0.841 & 0.222 & 0.690 & 1.0 \\ \hline
\multirow{2}{*}{COMPAS} & LR & 0.310 & 0.672 & 1.0 & 1.0 & 1.0 & 0.841 \\ \cline{2-8} 
 & NN & 0.151 & 1.0 & 1.0 & 0.690 & 1.0 & 0.548 \\ \hline
\end{tabular}
\caption{Consistency --- 1/36 significant}
\end{subtable}
\caption{P-values from Mann-Whitney U tests with null hypothesis that average metric values for group 0 ($m_0$) and group 1 ($m_1$) are equal. Of the 162 explanation/metric/model/dataset combinations analyzed, 18.5\% exhibited statistically significant explanation disparity (bolded). Note that the Ground Truth (Fidelity) metric can only be computed for linear models (LR).}
\vspace{-2em}
\label{pvalues}
\end{table*}

\begin{table*}[]
\small
\begin{tabular}{|c|c|c|c|c|c|c|c|}
\hline
\multicolumn{1}{|r|}{} &  & LIME & SHAP & SmoothGrad & IntGrad & VanillaGrad & Maple \\ \hline
\multirow{2}{*}{German Credit} & LR & 2 & 2 & 1 & 2 & 2 & 0 \\ \cline{2-8} 
 & NN & 0 & 0 & 1 & 1 & 1 & 0 \\ \hline
\multirow{2}{*}{Student Performance} & LR & 0 & 0 & 0 & 1 & 0 & 0 \\ \cline{2-8} 
 & NN & 0 & 2 & 0 & 2 & 0 & 1 \\ \hline
\multirow{2}{*}{COMPAS} & LR & 1 & 2 & 0 & 1 & 1 & 0 \\ \cline{2-8} 
 & NN & 0 & 2 & 1 & 2 & 2 & 0 \\ \hline
\end{tabular}
\caption{Here we aggregate the instances of significant explanation disparity reported in Table \ref{pvalues} across metrics, counting the number of times significant explanation disparity occurs in the explanation/model/dataset combinations.
Across these 36 combinations, explanation disparity occurs in at least one metric 56\% of the time.}
\vspace{-0.2in}
\label{counts}
\end{table*}

\subsection{Results \& Insights} 
We compute set averages $\overline{M_{0}}$ and $\overline{M_{1}}$ for each metric, model, explanation method, and dataset combination. To identify when explanation disparity is statistically significant, we perform Mann-Whitney U tests \cite{mannwhitney} on each pair of $\overline{M_{0}}$ and $\overline{M_{1}}$ distributions and report the resulting p-values in Table \ref{pvalues}. Of the 162 experimental combinations analyzed, 18.5\% exhibited statistically significant explanation disparity. In Table \ref{counts} we aggregate these instances across metrics, counting the number of times significant explanation disparity occurs in each of the 36 explanation, model, and dataset combinations. Across these 36 combinations, explanation disparity occurs in at least one metric 56\% of the time. Next we analyze explanation disparity across each of our framework metrics and summarize additional aggregate trends.

\subsubsection{Fidelity disparity}
\begin{figure*}[ht!]
\centering 
\begin{subfigure}{0.28\linewidth}
\centering
\includegraphics[width=\linewidth]{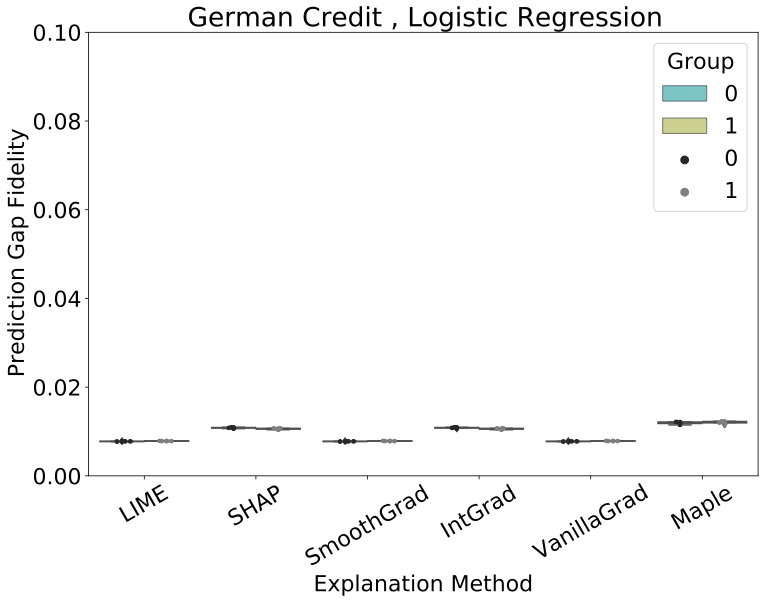}
\end{subfigure}
\begin{subfigure}{0.28\linewidth}
\centering
\includegraphics[width=\linewidth]{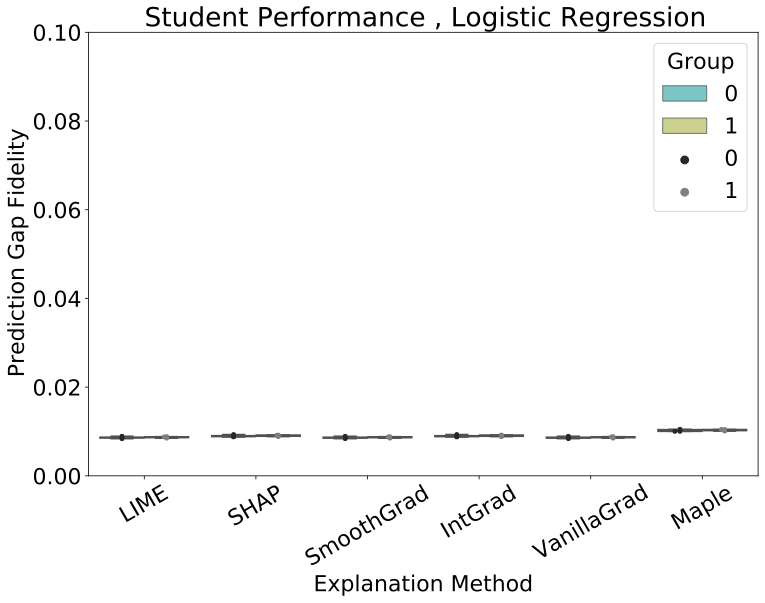}
\end{subfigure}
\begin{subfigure}{0.28\linewidth}
\centering
\includegraphics[width=\linewidth]{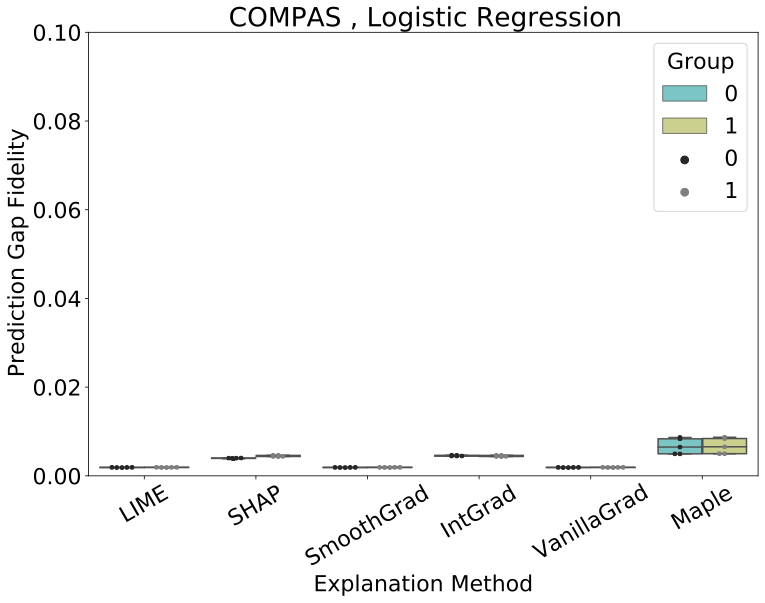}
\end{subfigure}
\begin{subfigure}{0.28\linewidth}
\centering
\includegraphics[width=\linewidth]{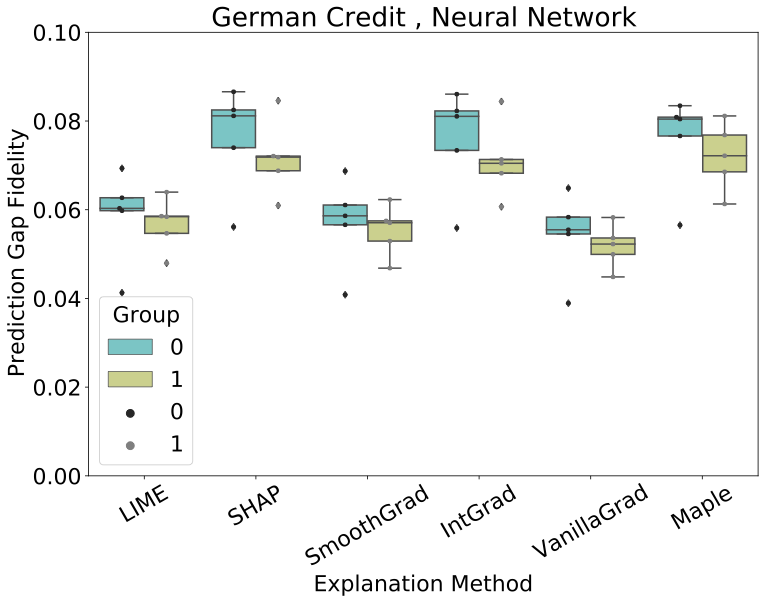}
\end{subfigure}
\begin{subfigure}{0.28\linewidth}
\centering
\includegraphics[width=\linewidth]{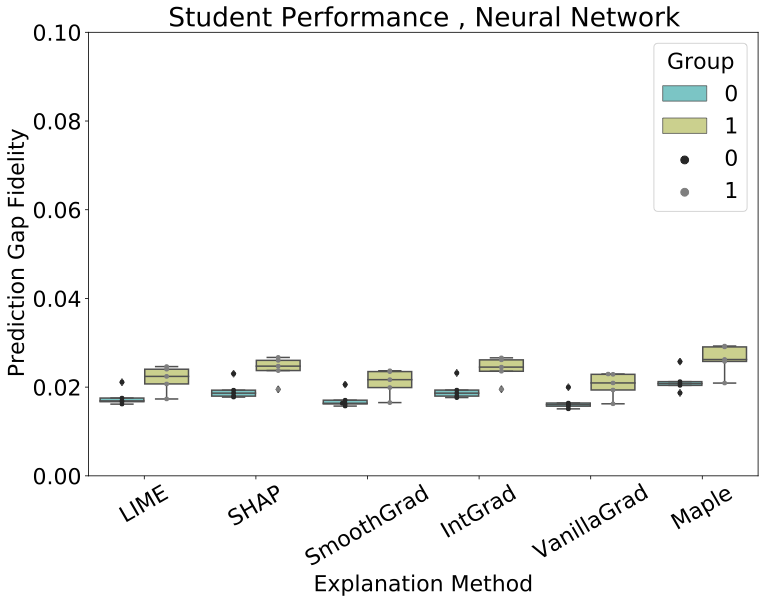}
\end{subfigure}
\begin{subfigure}{0.28\linewidth}
\centering
\includegraphics[width=\linewidth]{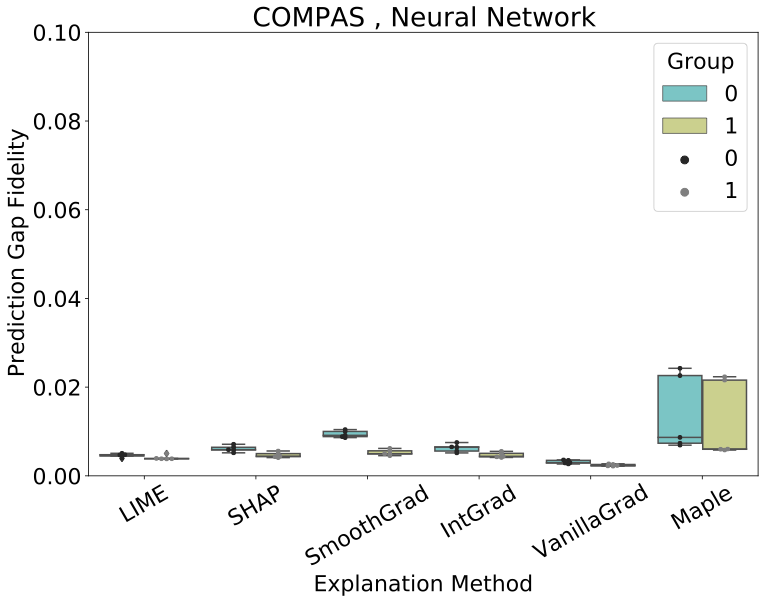}
\end{subfigure}
\caption{In each plot we compare the average prediction gap values for group 0 in blue ($m_0$) and group 1 in green ($m_1$) for each of the explanation methods ($\mathcal{E}$) listed on the x-axis. In the top row is German Credit (left), Student Performance (middle), COMPAS (right) with the LR predictive model. In the bottom row is German Credit (left), Student Performance (middle), COMPAS (right) with the NN predictive model. Differences between group 0 and group 1 distributions indicate explanation disparity. 
}
\label{predgapplots}
\end{figure*}

As detailed in subtables (a) and (b) of Table \ref{pvalues}, significant prediction gap fidelity disparity occurred 30.6\% of the time.
Significant ground truth fidelity disparity also occurred  11.1\% of the time.
Across both \textbf{ground truth fidelity} disparity and \textbf{prediction gap fidelity} disparity, significant fidelity disparity occurred most often with SHAP explanations.
For \textbf{prediction gap fidelity} disparity, which can be computed for both LR and NN models, first notice that the \textbf{prediction gap} values are larger across both groups 0 and 1 in the NN setting, indicating worse fidelity. Next notice that the majority of significant disparities occur in the more complex NN model setting (63.6\%). In addition to being more frequently significant, \textbf{prediction gap fidelity} explanation disparity appears more severe in complex model settings. This is illustrated in Figure \ref{predgapplots} by the wider gaps between $\overline{M_{0}}$ and $\overline{M_{1}}$ means in complex NN model settings when compared to their simpler LR counterparts. Moreover, even when disparities are not statistically significant, we notice more severe disparity in complex NN model settings when compared to their simpler LR counterparts. In these NN settings with observably larger disparity, the lack of statistical significance is possibly due to larger variances. These larger variances themselves suggest that prediction gap fidelity is a less robust metric on complex models. In general, the trends of fidelity disparity being more frequently significant and more severe in complex models align with our expectations of explanation methods that employ local linear approximations.

\subsubsection{Stability disparity}
\begin{figure*}[ht!]
\centering 
\begin{subfigure}{0.28\linewidth}
\centering
\includegraphics[width=\linewidth]{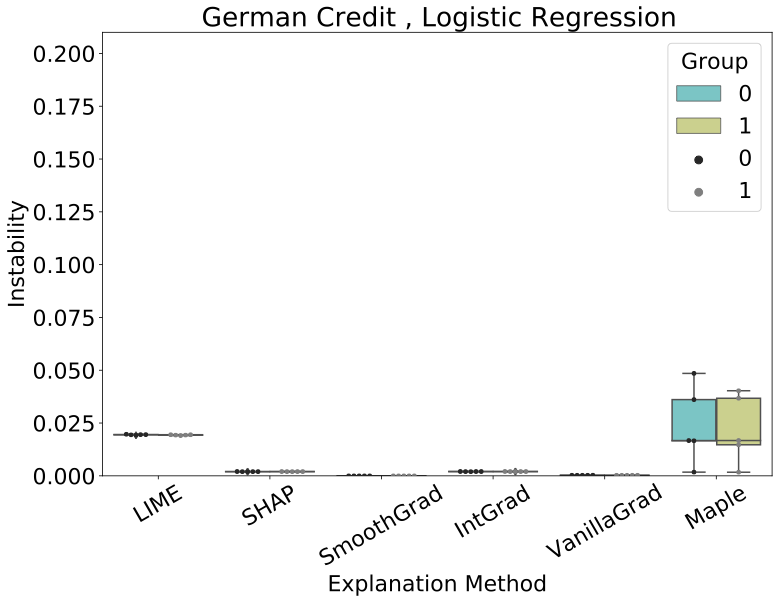}
\end{subfigure}
\begin{subfigure}{0.28\linewidth}
\centering
\includegraphics[width=\linewidth]{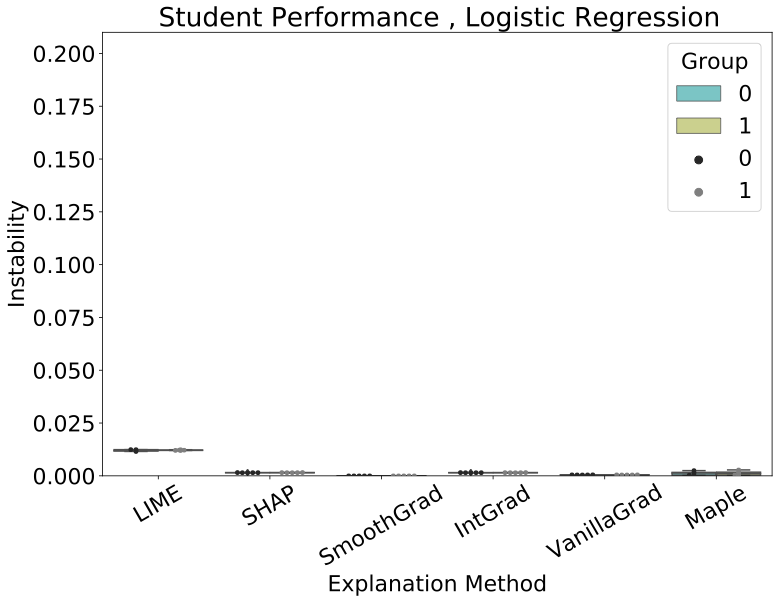}
\end{subfigure}
\begin{subfigure}{0.28\linewidth}
\centering
\includegraphics[width=\linewidth]{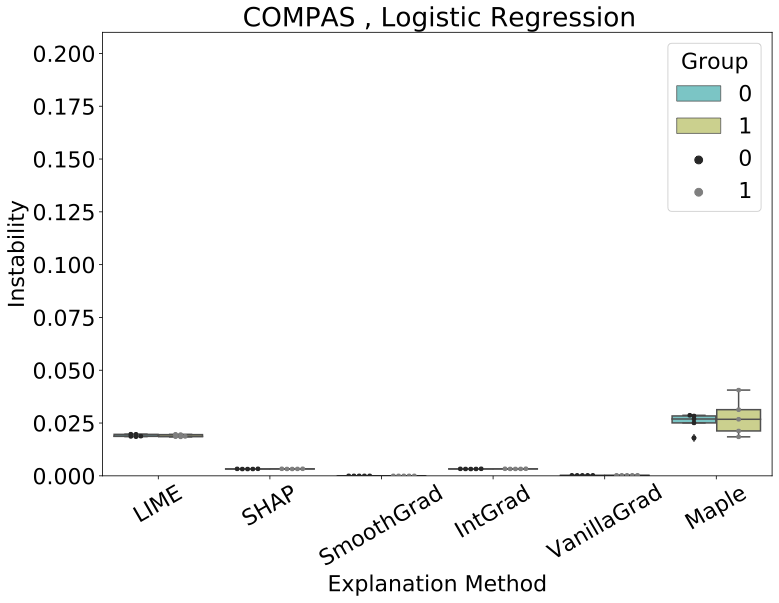}
\end{subfigure}
\begin{subfigure}{0.28\linewidth}
\centering
\includegraphics[width=\linewidth]{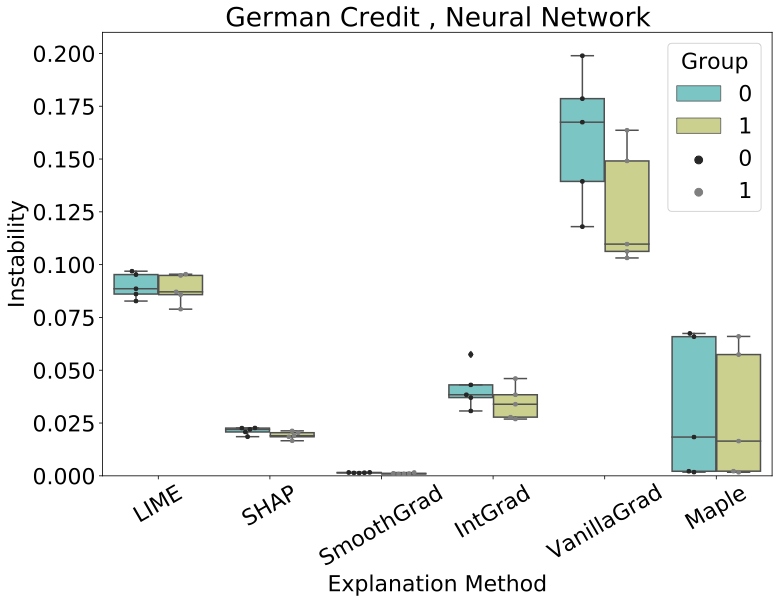}
\end{subfigure}
\begin{subfigure}{0.28\linewidth}
\centering
\includegraphics[width=\linewidth]{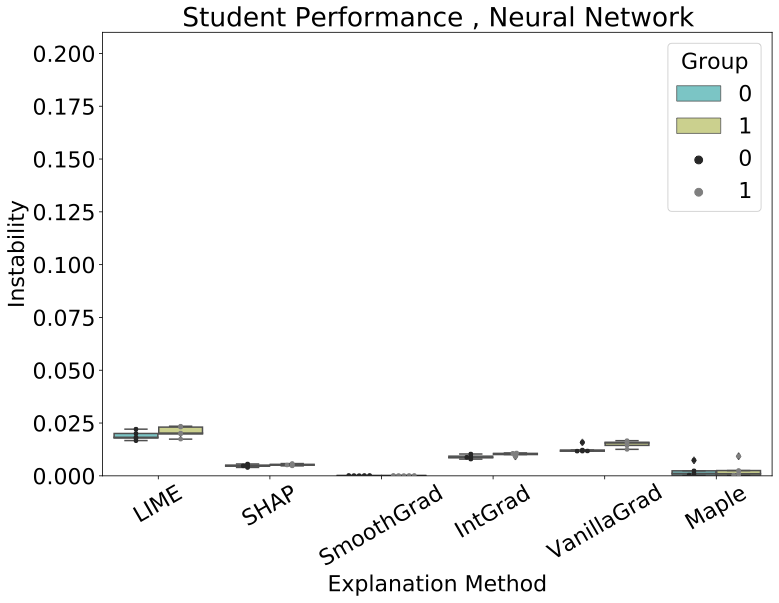}
\end{subfigure}
\begin{subfigure}{0.28\linewidth}
\centering
\includegraphics[width=\linewidth]{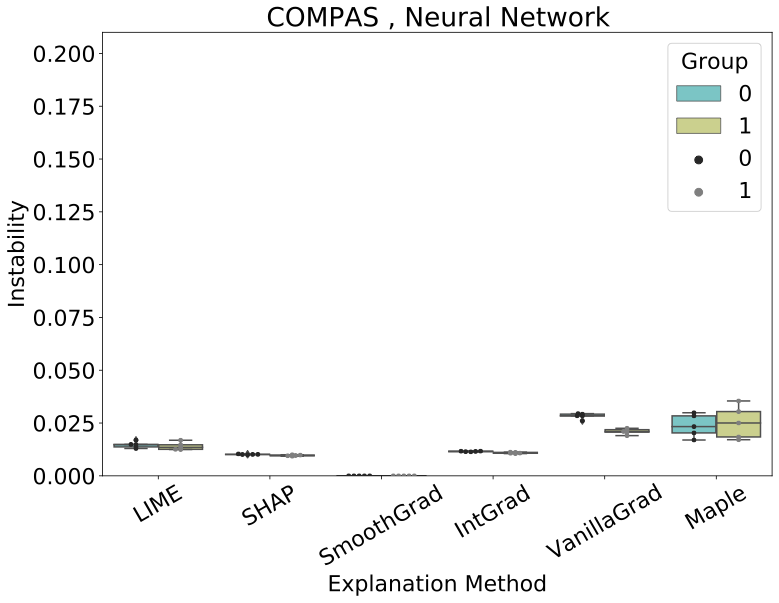}
\end{subfigure}
\caption{In each plot we compare the average stability values for group 0 in blue ($m_0$) and group 1 in green ($m_1$) for each of the explanation methods ($\mathcal{E}$) listed on the x-axis. In the top row is German Credit (left), Student Performance (middle), COMPAS (right) with the LR predictive model. In the bottom row is German Credit (left), Student Performance (middle), COMPAS (right) with the NN predictive model. Differences between the group 0 and group 1 distributions indicate explanation disparity. 
}
\label{robustnessplots}
\end{figure*}

As reported in Table \ref{pvalues} (d), significant \textbf{stability} disparity occurred 13.9\% of the time and most frequently with VanillaGrad explanations.
The majority (80\%) of instances of significant explanation disparity occurred on the complex NN model setting. In Figure \ref{robustnessplots}, we notice that instability in general, across both groups on the German Credit dataset. On the German Credit dataset, like with fidelity disparity, in Figure \ref{robustnessplots}, we notice that even when disparities are not significant, wider gaps between $\overline{M_{0}}$ and $\overline{M_{1}}$ means in complex NN model settings when compared to their simpler LR counterparts, indicating more severe disparity. Similarly, larger variances are again observed on more complex models, suggesting stability is a less robust metric in these settings.

\begin{figure*}[ht!]
\centering 
\begin{subfigure}{0.28\linewidth}
\centering
\includegraphics[width=\linewidth]{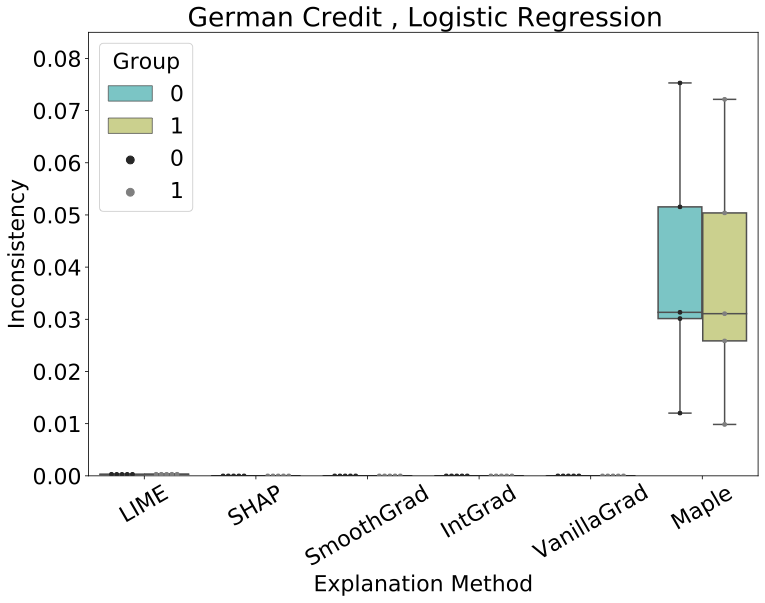}
\end{subfigure}
\begin{subfigure}{0.28\linewidth}
\centering
\includegraphics[width=\linewidth]{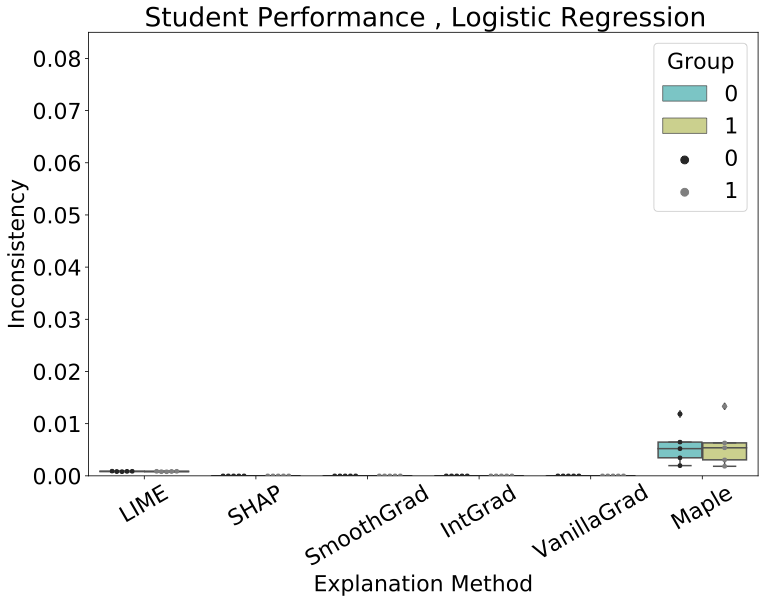}
\end{subfigure}
\begin{subfigure}{0.28\linewidth}
\centering
\includegraphics[width=\linewidth]{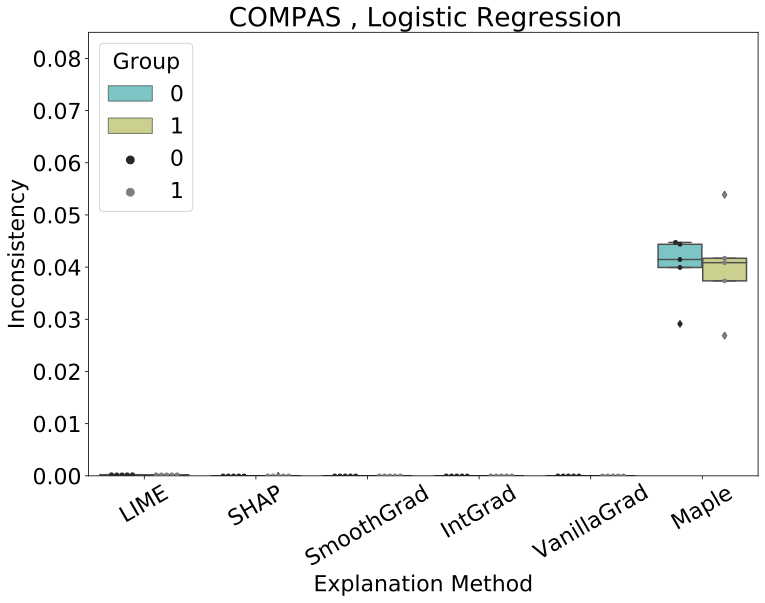}
\end{subfigure}
\begin{subfigure}{0.28\linewidth}
\centering
\includegraphics[width=\linewidth]{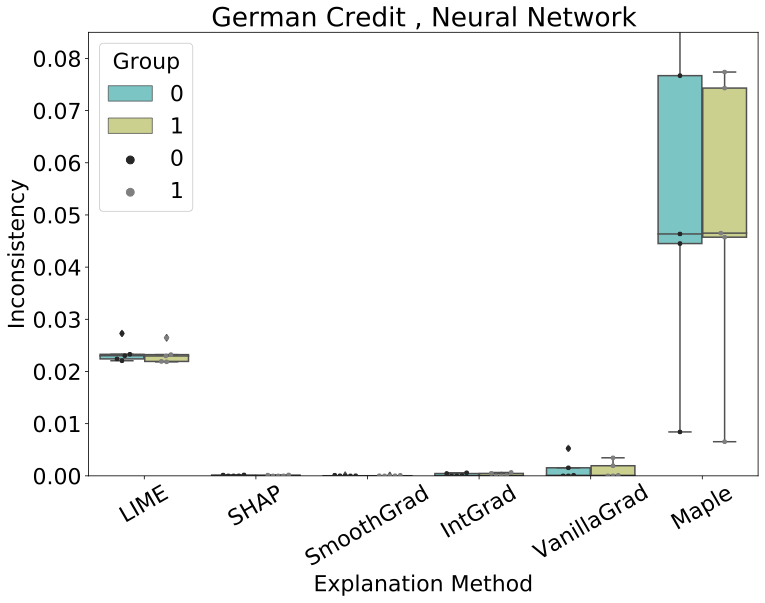}
\end{subfigure}
\begin{subfigure}{0.28\linewidth}
\centering
\includegraphics[width=\linewidth]{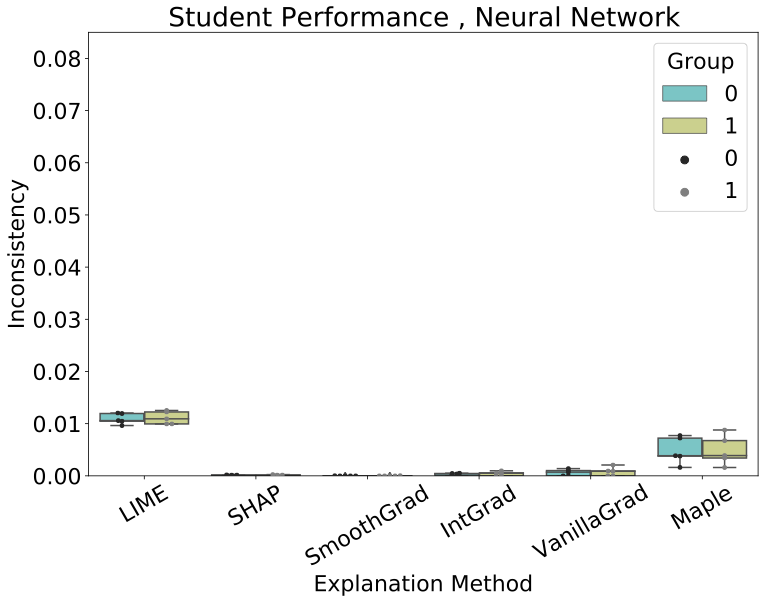}
\end{subfigure}
\begin{subfigure}{0.28\linewidth}
\centering
\includegraphics[width=\linewidth]{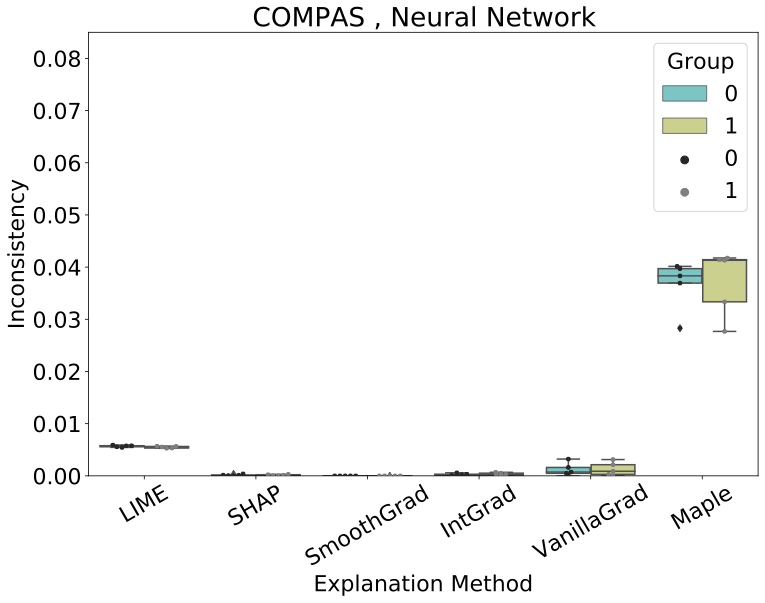}
\end{subfigure}
\caption{In each plot we compare the average consistency values for group 0 in blue ($m_0$) and group 1 in green ($m_1$) for each of the explanation methods ($\mathcal{E}$) listed on the x-axis. In the top row is German Credit (left), Student Performance (middle), COMPAS (right) with the LR predictive model. In the bottom row is German Credit (left), Student Performance (middle), COMPAS (right) with the NN predictive model. Differences between the group 0 and group 1 distributions indicate explanation disparity.}
\label{consistencyplots}
\end{figure*}

\begin{figure*}[h!]
\centering 
\begin{subfigure}{0.28\linewidth}
\centering
\includegraphics[width=\linewidth]{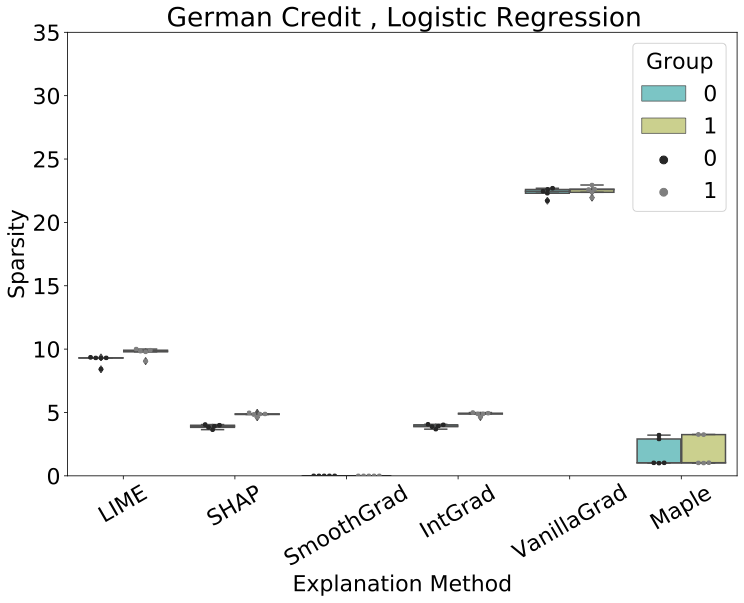}
\end{subfigure}
\begin{subfigure}{0.28\linewidth}
\centering
\includegraphics[width=\linewidth]{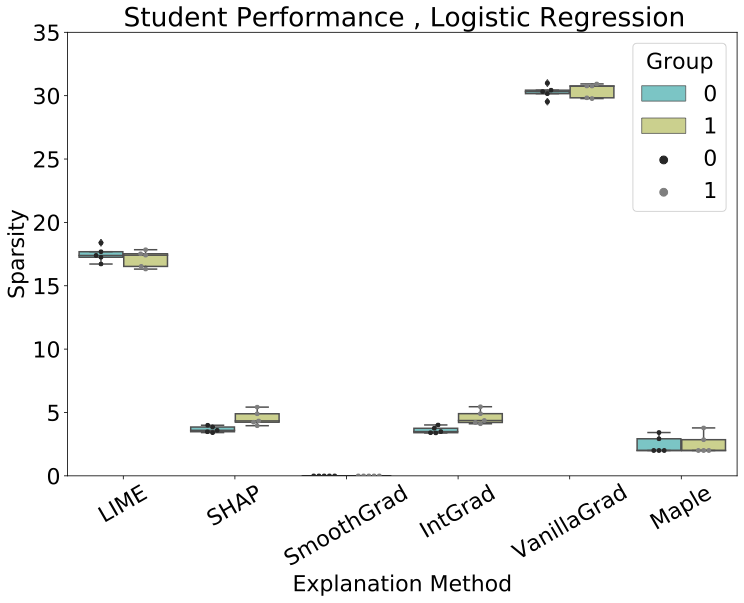}
\end{subfigure}
\begin{subfigure}{0.28\linewidth}
\centering
\includegraphics[width=\linewidth]{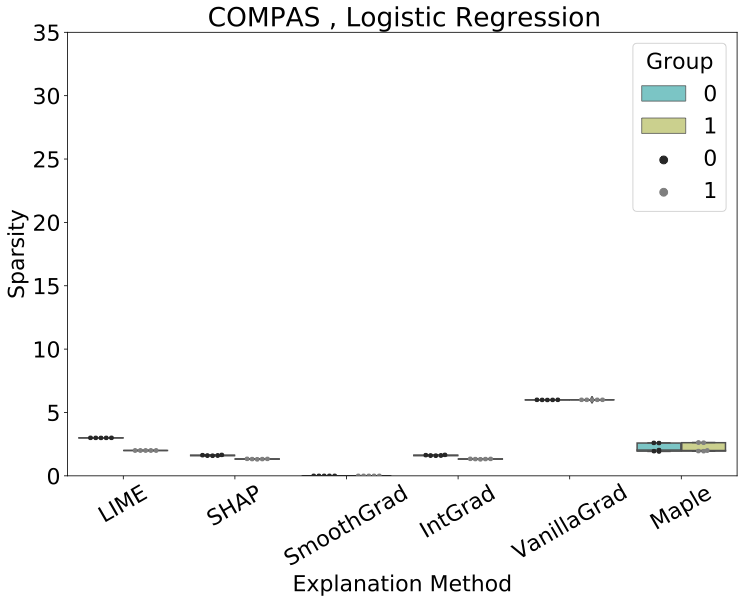}
\end{subfigure}
\begin{subfigure}{0.28\linewidth}
\centering
\includegraphics[width=\linewidth]{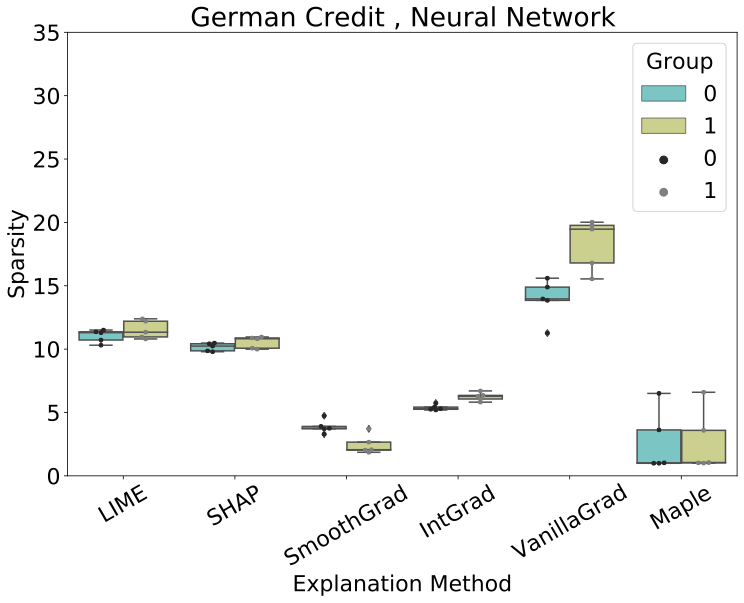}
\end{subfigure}
\begin{subfigure}{0.28\linewidth}
\centering
\includegraphics[width=\linewidth]{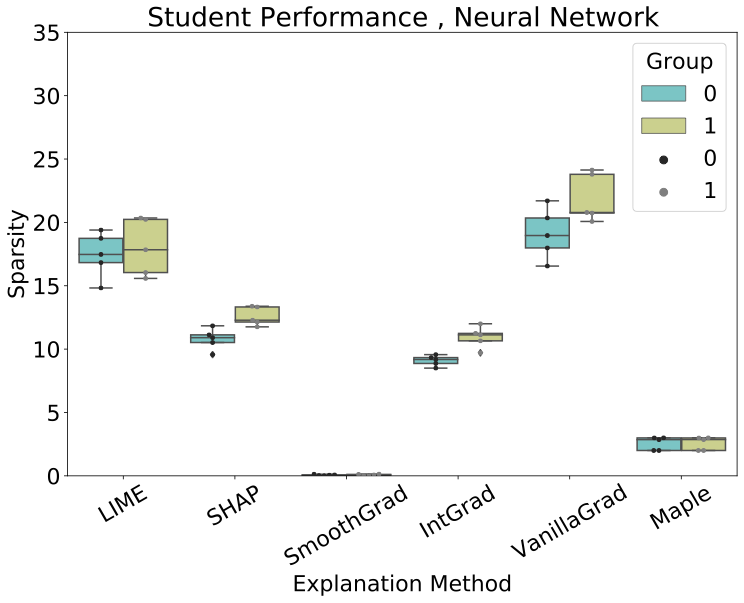}
\end{subfigure}
\begin{subfigure}{0.28\linewidth}
\centering
\includegraphics[width=\linewidth]{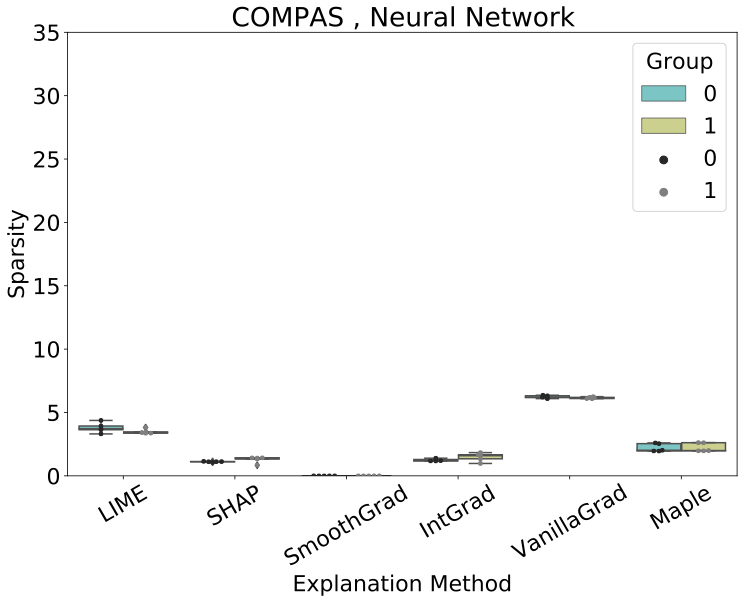}
\end{subfigure}
\caption{In each plot we compare the average sparsity values for group 0 in blue ($m_0$) and group 1 in green ($m_1$) for each of the explanation methods ($\mathcal{E}$) listed on the x-axis. In the top row is German Credit (left), Student Performance (middle), COMPAS (right) all with the LR predictive model. In the bottom row is German Credit (left), Student Performance (middle), COMPAS (right) all with the NN predictive model. Observe that differences between group 0 and and group 1 are present in both the LR and NN rows. Differences between the group 0 and group 1 distributions indicate explanation disparity.
}
\vspace{-0.2in}
\label{sparsityplots}
\end{figure*}

\subsubsection{Consistency disparity}
As reported in Table \ref{pvalues} (e), significant \textbf{consistency} disparity occurred in one setting, LIME on the LR model trained on the German Credit dataset. Yet as observed in Figure \ref{consistencyplots}, though there exists a statistically significant consistency disparity between group 0 and group 1 in this setting, it is not obvious because both groups 0 and 1 have very low inconsistency in general. In fact the only settings where inconsistency is often larger across both groups is with MAPLE explanations.

\subsubsection{Sparsity disparity}
As reported in Table \ref{pvalues} (c), \textbf{sparsity} disparity occurred 30.6\% of the time and most frequently with IntGrad explanations.
Notice that, in contrast with previously observed trends, instances of significant sparsity disparity are split more evenly between the NN and LR model settings, 45.5\% and 54.5\% respectively. However, in Figure \ref{sparsityplots} we notice that the variance in average sparsity is typically larger in NN settings than in LR counterparts, similarly to those for fidelity disparity, perhaps contributing to the fewer instances of statistically significant disparity.

\subsubsection{Additional aggregate trends}
To identify which explanation methods exhibit the most explanation disparity across our experiments, we read Table \ref{pvalues} columnwise, noting that each explanation method is employed on 27 metric, model, and dataset combinations. IntGrad exhibited significant explanation disparity on 33.3\% of the combinations, followed by SHAP (29.6\%), VanillaGrad (22.2\%), LIME (11.1\%), SmoothGrad (11.1\%), and MAPLE (3.7\%).
In Table \ref{counts} we also notice that IntGrad is the only explanation method with nonzero entries in every row. This means that IntGrad exhibits significant explanation disparity in at least one dimension in every model/dataset setting tested and is the only explanation method that does so.
As detailed in Section \ref{sec:discussion}, investigating the underlying mechanisms in explanation methods that give rise to these disparities is a critical direction for future research. 

To determine whether explanations disparity occurs across multiple metrics at once, we refer to Table \ref{counts}. Of the 20 explanation, model, and metric settings in Table \ref{counts} with non-zero entries, indicating significant explanation disparity is observed at least once, in one half disparities occur in only one metric, and in the other half disparities co-occur in two metrics.
This behavior suggests that practitioners should not rely on disparity in one metric to serve as a proxy for disparity along all metrics.

\section{Discussion}
\label{sec:discussion}

In this work, we proposed an evaluation framework for unearthing disparities in post hoc explanation quality. 
To the best of our knowledge, this work is the first to study the problem of group-based disparities across a variety of metrics for explanation quality.
This is especially important when the black-box model is making predictions about humans, given the consequential use of explanations in this context: explanations are often used in conjunction with a model in order to shape decisions, and disparity in explanation quality across demographic groups may be yet another way that adverse downstream outcomes become baked into the technical system. 
There are many cases where disparity in performance does \textit{not} occur. However, the prevalence of significant disparity throughout datasets, metrics, and methods in our results suggest that measuring disparity in explanation quality is still critical: more than half of all dataset-model combinations exhibited disparity.

Our work has lasting implications for several stakeholders. For researchers working on novel explanation methods, it may be useful to apply our framework on a test suite of datasets and models to identify whether the explanation method is susceptible to disparity \textit{in general}, and if so, with respect to what metrics. For practitioners who are developing application-specific models and explanation methods, our framework may be useful to identify whether the resulting explanations exhibit quality disparities in the context of the application. 

If disparity is in fact identified, what actions should stakeholders take? As discussed in Section \ref{sec:background}, disparity in fidelity and consistency always indicates that the explanation method is amplifying or generating problems. Disparity in stability and sparsity, on the other hand, may either indicate an issue with the explanation method, or an issue with the underlying model to be explained. 
Further research is necessary in order to provide more prescriptive next steps. Our work points to several follow-up research directions: First, what conditions give rise to observed disparity? Are there ways in which we can characterize datasets, models, or explanation methods such that we can more systematically identify the causes of measured explanation disparity? 
Second, human-grounded evaluations of our setting are necessary. Given significant explanation disparities, how is human decision making affected? 
Finally, it would be interesting to develop explanation methods which are less susceptible to such disparities while still maintaining overall utility.

\section*{Acknowledgments}
The authors would like to thank all the funding agencies supporting this work. This work is supported in part by the NSF awards IIS-2008461 and IIS-2040989, 
and research awards from Amazon, Harvard Data Science Institute, Bayer, and Google. HL would also like to thank Sujatha and Mohan Lakkaraju for their support and encouragement. 
The views expressed here are those of the authors and do not reflect the official policy or position of the funding agencies. 
JD completed this work while an undergraduate at Brown University.

\bibliographystyle{ACM-Reference-Format}
\bibliography{refs.bib}


\end{document}